%% file: acl_latex.tex
\pdfoutput=1
\documentclass[11pt]{article}

\usepackage[final]{acl}

\usepackage{times}
\usepackage{latexsym}

\usepackage[T1]{fontenc}
\usepackage[utf8]{inputenc}

\usepackage{microtype}

\usepackage{inconsolata}

\usepackage{graphicx}
\usepackage{cleveref}
\usepackage{booktabs}
\usepackage{subcaption}
\usepackage{array}

\usepackage{caption}
\usepackage{subcaption}

\usepackage{booktabs}
\usepackage{multirow}
\usepackage{makecell}
\usepackage{arydshln}

\usepackage{algorithm} 
\usepackage{algpseudocode}
\usepackage{microtype}

\usepackage{graphicx}
\usepackage{booktabs}
\usepackage{utility/dm_utility}

\usepackage{hyperref}
\usepackage{enumitem}
\usepackage[title]{appendix}
\usepackage{soul}
\usepackage[skins]{tcolorbox}
\usepackage{graphicx}
\usepackage{enumitem}
\usepackage{makecell}
\tcbuselibrary{listingsutf8}
\global

\tcbset{
  aibox/.style={
    width=474.18663pt,
    top=10pt,
    colback=white,
    colframe=black,
    colbacktitle=black,
    enhanced,
    center,
    attach boxed title to top left={yshift=-0.1in,xshift=0.15in},
    boxed title style={boxrule=0pt,colframe=white,},
  }
}
\newtcolorbox{AIbox}[2][]{aibox,title=#2,#1}

\title{What's Wrong? Refining Meeting Summaries with LLM Feedback}

\author{Frederic Kirstein\textsuperscript{1,2}, Terry Ruas\textsuperscript{1}, Bela Gipp\textsuperscript{1} \\
  \textsuperscript{1}Georg-August-Universität Göttingen, Germany \\
\textsuperscript{2}\texttt{kirstein@gipplab.org} }

\begin{document}
\maketitle
\begin{abstract}
Meeting summarization has become a critical task since digital encounters have become a common practice.
% Lately, many summarization assistants have been introduced, often based on large language models (LLMs). 
Large language models (LLMs) show great potential in summarization, offering enhanced coherence and context understanding compared to traditional methods.
However, they still struggle to maintain relevance and avoid hallucination.
We introduce a multi-LLM correction approach for meeting summarization using a two-phase process that mimics the human review process: \textbf{mistake identification} and \textbf{summary refinement}.
We release QMSum Mistake, a dataset of 200 automatically generated meeting summaries annotated by humans on nine error types, including structural, omission, and irrelevance errors.
Our experiments show that these errors can be identified with high accuracy by an LLM.
We transform identified mistakes into actionable feedback to improve the quality of a given summary measured by relevance, informativeness, conciseness, and coherence.
This post-hoc refinement effectively improves summary quality by leveraging multiple LLMs to validate output quality.
Our multi-LLM approach for meeting summarization shows potential for similar complex text generation tasks requiring robustness, action planning, and discussion towards a goal.
\end{abstract}

\section{Introduction}
\input{text/01_introduction}

\input{tables/03_dataset/QMSum_overview}

\section{Related Work}
\input{text/02_relatedwork}

\section{QMSum Mistake Dataset}
\input{text/03_dataset}

\section{Mistake Identification}
\label{sec:MistakeIdentification}

\input{text/04_error_identification}

\section{Summary Refinement}
\label{sec:SummaryRefinement}
\input{text/05_refinement}

%\section{Multi-agent Discussion}
%\input{text/06_discussion_paradigm}

%\section{Scenarios or Ablation}
%\input{text/07_scenarios}

% \section{Discussion}
\section{Final Considerations}
In this paper, we investigated GPT4's ability to find mistakes in a given meeting summary and refine them accordingly.
We found that GPT4 achieves a high accuracy of $\sim$89\% on average, measured against human labels, in identifying typical mistakes (e.g., repetition of content) when using a dedicated model instance paired with CoT prompting to identify individual errors.
However, it struggles to identify similar and subjective errors, such as hallucination (72\% acc.) with omission and irrelevance (81\% acc.).
We showed strong evidence that a dedicated LLM can refine a summary based on identified errors.
By providing a CoT explanation for each error type containing reasoning why and where an error was observed, we improve the quality of relevance, informativeness, conciseness, and coherence significantly.
These refined summaries are comparable in quality, with error-free gold summaries.
Our post hoc refinement approach can be applied to refine meeting summaries generated by traditional models and LLMs and marks an early entry into methods that allow the full potential of LLMs for meeting summarization.
We leave the development of more sophisticated refinement protocols, e.g., using multi-agent discussion, and the application of our multi-LLM approach to similar complex text generation tasks (e.g., story writing to reflect on given setting) and real-world applications (e.g., assisting LLM agents to check the outcome to a task) to future work.
We release QMSum Mistake to encourage research on refinement.

\section*{Acknowledgements}
This work was supported by the Lower Saxony Ministry of Science and Culture and the VW Foundation.
Frederic Kirstein was supported by the Mercedes-Benz AG Research and Development.

\section*{Potential Impact}
The multi-LLM approach proposed here, influenced by psychological observations on productivity and collaboration, exemplifies how other academic fields can inform NLP research \cite{WahleRAG23}.
This work demonstrates the potential for enhancing complex text generation tasks requiring robust output such as machine translation \cite{FengZLW24}, reasoning \cite{KalyanpurSBC24}, question answering \cite{KimKY24}, or paraphrasing \cite{BeckerWRG23,WahleGR23}, that may benefit from an output-challenging system that assesses content alignment.
By incorporating multi-LLM strategies and personalization, we open new avenues for improving NLP outputs across various applications, underscoring the value of interdisciplinary approaches in advancing NLP technologies and their real-world applicability.

\section*{Limitations}
% One main limitation of our dataset is its comparable small size of 200 samples in total.
% As the annotation of the erroneous samples required significant time (total of $\sim$ 1000h), we decided to limit the number here.
% However, to provide a meaningful dataset, we took special care to ensure the samples are representative of each language model, that the number of samples is significant (measured against the original size of QMSum), and that it presents a balanced representation of the different sub-dataset (AMI, ICSI, parliament).
% We further took care that sub-dataset-related patterns can be analyzed by including automatic summaries that share the same transcript but use different language models.
Although our proposed QMSum Mistake might seem small (i.e., 200 samples), its size is comparable to the original QMSum dataset (i.e., 232 samples).
We contribute to extending the original dataset with careful human error annotations for almost all examples available.
Another possible limitation in our work is the use of only GPT4 in our main experiments.
We chose GPT4 because of its large context size (e.g., 128k tokens) and better initial results in identifying errors.
Evaluating and error annotation and refinement for multiple models by humans would be time-consuming and financially unfeasible.
However, we report the detailed results in \Cref{sec:use_case} to provide insights on other language families and different models (e.g., Phi \cite{AbdinJAA24}, Gemini \cite{GeminiTeamRST24}) considered in our study.
We evaluate their performance on mistake identification and quality changes when refining a summary.
% We excluded LLMs with a context size of less than 16k tokens from this comparison (e.g., Llama 3) as this would require major changes in the overall architecture of our refinement system, which leads to limited comparability.

% Another limitation is that we use GPT-4 Turbo to generate refined summaries and for the evaluation and ranking to compare the quality against non-GPT-4 Turbo summaries.
% This may induce a bias of the model to favor text generated by GPT-4 Turbo over others.
% We chose an LLM-based evaluation method for scalability, so we could analyze all generated summaries and show initial evidence of the refinement performance.
% Further, our main focus is to compare the quality changes when applying refinement, hence comparing GPT-4 Turbo generated texts against each other.
% To nevertheless mitigate this error, we exhaustively tested the different LLM-based evaluation methods and checked a significant portion of the scores reported in this work with the help of our human annotators to ensure that the scores align closely with human annotations.

\section*{Ethics Statement and Broader Impact}
Our research abides by ethical guidelines for AI research and is committed to privacy, confidentiality, and intellectual property rights.
We've ensured that the datasets in our study, publicly available, do not house sensitive or personal details.
While our study leverages existing resources and generative models, it's important to note that these models can possess biases and may occasionally generate summaries with distortions, biases, or inappropriate content.
To counteract this, we've configured our models to omit potentially harmful or unsafe content.
While our research aims to enhance meeting summarization to benefit communication and productivity across sectors, we're acutely aware of the ethical challenges posed by AI in this domain.
Meeting summarization models must be wielded with respect to privacy and consent, especially when processing sensitive or confidential material.
It's paramount that these models neither violate privacy nor perpetuate harmful biases.
As the field evolves, we stress the importance of maintaining these ethical considerations and encourage fellow researchers to uphold them, ensuring that AI advancements in meeting summarization are both beneficial and ethically grounded.
An integral aspect of our ethical commitment is reflected in our approach to annotator recruitment and management.
The team of annotators, consisting of interns, student assistants, and doctoral students, was meticulously selected through internal channels.
This strategy was chosen to uphold a high standard of annotation quality—a quality we found challenging to guarantee through external platforms such as Amazon Mechanical Turk.
Ensuring fair compensation, these annotators were remunerated in accordance with institutional guidelines for their respective positions.
Further, flexibility in the annotation process was also a priority.
Annotators had the freedom to choose their working times and environments to prevent fatigue from affecting their judgment.

% This work was partially supported by the Lower Saxony Ministry of Science and Culture, and the VW Foundation. Many thanks to ... for their thoughtful discussions.
%TR please make sure to add :https://gipplab.atlassian.net/wiki/spaces/ISG/pages/57748546/Acknowledge+Funding+in+Publications  - Because of me and Bela we need to acknowledg the VW foundation. Maybe you can also add something from the MB side? The annotators, or the division that provided them, etc :)

\bibliography{24_EMNLP_FeedbackAgent, impact}

\appendix

%%%%%%%%%%%%%%%%%%%%
%%%
\section{Human Annotation}
\input{text/A_HumanAnnotation}

\section{Exploring Additional Model Families and Setups}
\label{sec:use_case}
\input{text/B_AdditionalModelFamilies}

\section{QMSum Mistake varying summarization styles and quality levels of models}
\input{text/C_QMSum_Mistake_Examples}

\section{Prompts}
\input{text/D_Prompts}

\section{Additional Content on Summary Refinement}
\subsection{Established Metrics' scores}
\input{text/E_AdditionalContentRefinement}

\subsection{Correction and CoT are contradictory }
\input{text/F_Correction_CoT_contradiction}

\end{document}

%% file: text/01_introduction.tex
\input{figures/callables/main_figure}
Meeting summaries are essential for professional conversations, they serve as a reference for subsequent processes, update absentees, and reinforce the most important topics discussed.
The growing importance of summarization systems is evident from the recent release of tools in virtual meeting software (e.g., Zoom\footnote{\href{https://www.zoom.com/en/ai-assistant}{https://www.zoom.com/en/ai-assistant}},
Microsoft Teams\footnote{\href{https://copilot.cloud.microsoft}{https://copilot.cloud.microsoft}}, 
Google Meet\footnote{\href{https://support.google.com/meet/answer/14754931}{https://support.google.com/meet/}}).
Still, meeting summarization faces challenges, such as handling spoken language idiosyncrasies and identifying salient content \cite{KirsteinWGR24}.
Existing techniques, like AMR-graphs for capturing speaker relations \cite{HuaDM23a}, are often tailored to specific backbone models, typically using BART \cite{LewisLGG20b}, PEGASUS  \cite{ZhangZSL20a} or their variations.
Recent explorations of large language models (LLMs) for meeting summarization reveal their strong capabilities (e.g., high-quality summaries of long inputs) \cite{LaskarFCB23}.
However, these LLM-generated summaries are still error-prone \cite{KirsteinWRG24} and costly to fine-tune \cite{ChauhanRDG22a,WangZZC22d}.
%Existing methods may not be suitable for turning LLMs into robust and reliable meeting summarizers, as the techniques often rely on costly fine-tuning \cite{ChauhanRDG22a,WangZZC22d}. % and may become outdated with new LLM capabilities.

The shift to LLMs as backbone models raises the question of how to better use their capabilities and mitigate their weaknesses.
(Self-)correction through few-shot prompting improves LLM performance by asking it to review and correct its output \cite{PanSXN23}.  % MiaoTR23
While successful in various tasks (e.g., question answering \cite{JiangZWW24}, reasoning \cite{MadaanTRC21}, and summarization \cite{SaundersYWB22}), self-correction still falls short to identify and correct errors \cite{HuangCMZ24}.
To address this, \citet{TyenMCC24} propose a multi-LLM refinement process for reasoning tasks with, leading to a more robust correction approach.

Analogous to how humans iterate over suggestions and edits when writing texts, we explore how LLMs may be employed in the same way to improve meeting summarization in a two-stage approach consisting of \textbf{mistake identification} in an existing summary and a subsequent \textbf{refinement} (\Cref{fig:overview_setup}).
For mistake identification, we annotate QMSum \cite{ZhongYYZ21f} on nine error types (e.g., omission, structural mistakes) \cite{KirsteinWRG24,ChangLGI24}.
GPT-4 Turbo\footnote{We will refer to this as GPT4 throughout the paper.} identifies errors on average with $\sim$89\% accuracy, but it struggles with irrelevance ($\sim$81\%) and hallucination ($\sim$72\%) errors.
We achieve the best results on the mistake identification task using multiple LLM instances for each error type and Chain-of-Thought (CoT) prompting \cite{WeiWSB23b}.
For the refinement stage, we use an additional model instance to adjust an erroneous summary according to the detailed feedback from the mistake identification stage.
We explore what content a refinement model requires, considering the CoT explanation from the mistake identification task, a correction suggestion, and the original meeting transcript as additional information sources for pointed-out mistakes.
We further analyze if the feedback should be passed through an intermediate planning stage that extracts which content to add, remove, or rewrite in a summary.
We identify strong quality improvements for refined summaries over the original ones and baselines when using the CoT explanation from the mistake identification as feedback along the erroneous summary without additional processing.
Our contributions are summarized as follows:

\begin{itemize}[parsep=0.3pt,itemsep=0.1pt]
    \item QMSum Mistake\footnote{The dataset will be later available through Huggingface and the project-accompanying Github repository.}, a dataset of 200 meeting summaries and human-annotated errors. 
    \item A multi-LLM approach to finding mistakes in meeting summaries considering different prompting approaches.
    \item A transformation of identified mistakes into actionable feedback to refine an erroneous summary and derive a refinement protocol.
\end{itemize}

%% file: figures/callables/main_figure.tex
\begin{figure*}[t]
  \includegraphics[width=1\textwidth]{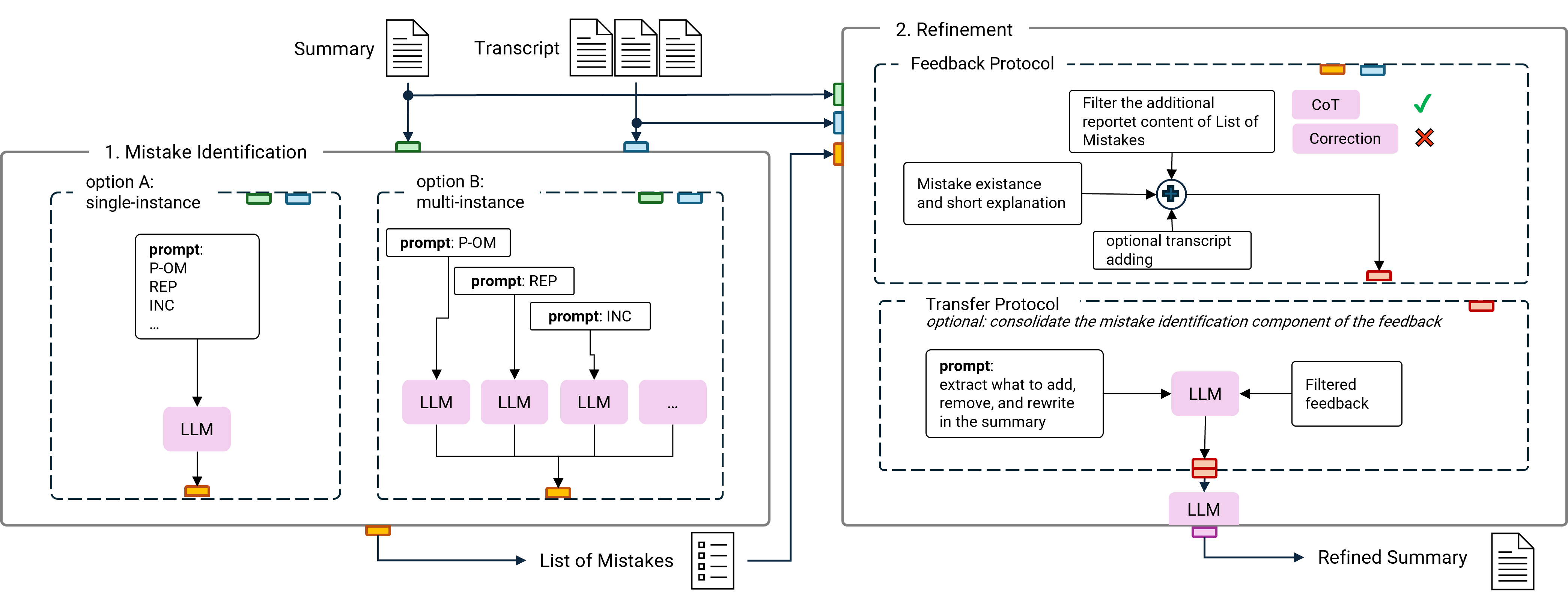}
  \caption {Overview of the two-stage refinement protocol displaying the assessed variants. The Mistake Identification block is analyzed \Cref{sec:MistakeIdentification} and the Refinement block in \Cref{sec:SummaryRefinement}.}
   \label{fig:overview_setup}
\end{figure*}

%% file: tables/03_dataset/QMSum_overview.tex
\begin{table*}
  \small
  \centering
    \begin{tabular}{lcccccc}
    \toprule
    Dataset & \# Meetings & \# Turns & \# Speakers & \# Len. of Meet. & \# Len. of Gold Sum. & \# Len. of Aut. Sum. \\
    \midrule
    AMI  &  124 (113)  & 535.6 & 4.0   & 6007.7 & 108.8 & 112.4 \\
    ICSI   & 52 (42)  & 819.0   & 6.3   & 13317.3 & 103.0 & 108.2 \\
    WPCP  & 24 (14)  & 207.7 & 34.1  & 13761.9 & 129.5 & 112.9 \\
    \midrule
    QMSum Mistake & 200 (169)   & 556.8 & 9.2   & 9069.8 & 109.1 & 116.9 \\
    \bottomrule
    \end{tabular}%
  \caption{Statistics for the QMSum Mistake dataset. Values are averages of the respective categories. Lengths (Len.) are in number of words. In \# Meetings, values in parentheses are the number of erroneous samples.}
  \label{tab:statistics_QMSum}%
\end{table*}%

%% file: text/02_relatedwork.tex
%%%%%%%%%%%%%%%%%%%%%%%%%%%%%%%%%%%%%%%%%%%%%%%%%%%%%%%%%%%%%%%%%
%% @TR This part did not change since you reviewed it last time!
%TR -tks, skipping to End
%%%%%%%%%%%%%%%%%%%%%%%%%%%%%%%%%%%%%%%%%%%%%%%%%%%%%%%%%%%%%%%%%

\noindent \textbf{Meeting Summarization} and its parent domain dialogue summarization are transitioning from traditional encoder-decoder models to LLMs.
Traditional models, such as BART \cite{LewisLGG20b} and PEGASUS \cite{ZhangZSL20a}, improved through techniques tailored to specific challenges like language, structure, comprehension, speaker, salience, and factuality \cite{KirsteinWGR24, KirsteinWRG24}.
These models integrated methods such as AMR-graphs for speaker relations \cite{HuaDM23a}, role vectors for speaker correlation \cite{AsiWEG22a,NarakiSH22a}, and additional training stages to bridge the gap between pre-training on written texts and spoken dialogue tasks \cite{RaffelSRL20a,KhalifaBM21b,LeeYPS21a}.
Recently, LLMs have been explored for meeting summarization by prompting the model to create a TL;DR \cite{LaskarFCB23,KirsteinWRG24}, showing comparable performance to specialized encoder-decoder models but with better context comprehension.
They thereby use LLMs without any adaptations and serve as the first works to report on LLM performance on meeting summarization.
Our work examines the effectiveness of LLMs as post-processors for summaries, assessing if this approach can achieve high-quality summaries without requiring techniques tailored to a specific challenge of meeting summarization.
We compare this against original summaries, single-LLM baselines, and human summaries, providing an updated benchmark for LLMs in meeting summarization.
For the creation of QMSum Mistake, we extend the work by \citet{KirsteinWRG24}, refining their definition of errors.
% Further, their analysis of evaluation metrics informed our choice in \Cref{sec:refinement-evaluation}.

\noindent \textbf{Self-correction} methods have been extensively studied in recent literature \cite{PanSXN23}, including training-time correction strategies like Reinforcement Learning from Human Feedback (RLHF) \cite{OuyangWJA22} and self-improvement techniques \cite{HuangCMZ24}.
Our feedback and refinement method falls into the category of post-hoc correction, which is applied to outputs already generated.
Previous post-hoc correction methods, such as Reflexion \cite{ShinnCBG23a} and RCI \cite{KimBM23}, focus on reasoning errors and often degrade performance without oracle labels \cite{HuangCMZ24}.
Our work uniquely applies post-processing correction to meeting summarization, focusing on qualitative improvements with independent models, and further explores this to other model families and related summarization domains.
Our approach is informed by the two-stage setup of \cite{TyenMCC24} which we extend with an extensive mistake identification architecture and a multi-stage refinement.

%% file: text/03_dataset.tex
% \begin{table*}
%   \centering
%   \small
%   \begin{tabular}{lc}
%     \toprule
%     Gold Summary   & ABC           \\
%     \midrule 
%     Automatic Summary & ABC           \\
%     \midrule
%     Omission & 1.0 | REASONING \\
%     Repetition & 1.0 | REASONING \\
%     Incoherence  & 1.0  | REASONING \\
%     Coreference   & 1.0 | REASONING \\
%     Hallucination (extrinsic)  & 1.0 | REASONING \\
%     Linguistic Inaccuracy (intrinsic)     & 1.0 | REASONING \\
%     Misrepresentation of Structure    & 1.0 | REASONING \\
%     Irrelevance & 1.0 | REASONING \\
%     \bottomrule
%   \end{tabular}
%   \caption{Example of an automatic summary generated by DIALOGLED. The human rating on the examples and the corresponding reasoning is stated below.}
%   \label{tab:Example_Annotation}
% \end{table*}

QMSum Mistake consists of 200 samples, with 169 (85\%) automatically created meeting summaries annotated on nine error types (\Cref{sec:errors}) and 31 error-free summaries serving as controls to analyze if the mistake identification is too sensitive.
\Cref{tab:statistics_QMSum} provides dataset statistics.
The samples stem from QMSum's \cite{ZhongYYZ21f} training and test sets, including AMI (staged business meetings) \cite{CarlettaKAB05}, ICSI (academic meetings) \cite{JaninBEE03}, and parliament meetings.
As gold summaries lack typical errors of automatic summaries, we generate summaries using encoder-decoder models (i.e., LED \cite{BeltagyPC20c}, DialogLED \cite{ZhongLXZ22f}, PEGASUS-X \cite{PhangZL22c}) for more severe mistakes in automatic summaries such as coreference and structure errors and LLMs (i.e., GPT-3.5, Phi-3 mini 128k \cite{AbdinJAA24}) for subtle errors such as relevance.
Models have a context size of at least 16k to fit the entire meeting in the input, use default settings, and generate up to 200 tokens to match gold summary lengths.
\Cref{tab:qmsum_mistake_examples} shows examples of varying summarization styles and quality levels.
The human annotation process, which achieved an average Krippendorff's alpha of 0.780 (see \Cref{tab:krippendorffs_alpha_human_annotation}), is described in \Cref{sec:prompts}.

\subsection{Observable errors}
\label{sec:errors}
\input{tables/03_dataset/error_definitions}

We refine existing error types \cite{KirsteinWRG24,ChangLGI24} into nine error types with minimal overlap.
\Cref{tab:error_definition} holds the short definitions.
Preliminary testing and annotator feedback inform the refinement of the error types and point out overlap in error definitions, making a clear distinction difficult.
This leads to major adaptations to precisely delimit the repetition, incoherence, structure, and linguistic inaccuracy errors, while the omission errors undergo minor tweaks in wording.
Hallucination errors are packed into a single category to reduce overlap for edge cases between these two.
The initial observations further indicate that errors so far were designed to capture missing or incorrect information, not the inclusion of unrelated content, which our summary-generating models tend to generate.
Thus, we add the 'Irrelevance' category.

%% file: tables/03_dataset/error_definitions.tex
\begin{table*}
\centering
\small
\begin{tabular}{p{2cm} p{2cm} p{10.5cm}}
\toprule
\textbf{Error Type} & \textbf{Transcript} & \textbf{Definition} \\
\midrule
Redundancy \newline \textbf{RED} & not required & The summary contains repeated or redundant information, which does not help the understanding or contextualization. \\
\midrule
Incoherence \newline \textbf{INC} & not required & The model generates summaries containing characteristics that disrupt the logical flow, relevance, or clarity of content either within a sentence (intra-sentence) or across sentences (inter-sentence). \\
\midrule
Language \newline \textbf{LAN} & not required & The model uses inappropriate, incorrect (ungrammatical), or ambiguous language or fails to capture unique linguistic styles. \\
\midrule
Omission \newline (partial, total) \newline \textbf{P-OM}, \textbf{T-OM} & required & Missing information from the meeting, such as significant decisions or actions. \textbf{Total omission:} Relevant topics and key points are not stated. \textbf{Partial omission:} Salient topics are mentioned but not captured in detail. \\
\midrule
Coreference \newline \textbf{COR} & required & The model fails to resolve a reference to a participant or entity, misattributes statements, or omits necessary mentions. \\
\midrule
Hallucination \newline \textbf{HAL} & required & The model produces inconsistencies not aligned with the meeting content. \textbf{Intrinsic:} Misrepresents information from the transcript. \textbf{Extrinsic:} Introduces content not present in the transcript. \\
\midrule
Structure \newline \textbf{STR} & required & The model misrepresents the order or logic of the meeting's discourse, misplacing topics or events. \\
\midrule
Irrelevance \newline \textbf{IRR} & required & The summary includes information that is unrelated or not central to the main topics or objectives of the meeting. \\
\bottomrule
\end{tabular}
\caption{Definition of the nine error types annotated in QMSum Mistake based on existing error types \cite{KirsteinWRG24,ChangLGI24}}
\label{tab:error_definition}
\end{table*}

%% file: text/04_error_identification.tex
\Cref{tab:breakdown_feedback_identification} shows GPT4's\footnote{gpt-4-turbo-2024-04-09, default settings, temperature = 0} accuracy in identifying summarization-related errors (\Cref{sec:errors}) on the QMSum Mistake dataset.
We chose GPT4 for its context size, understanding capabilities, robustness to handle spoken language, and superior results compared to Gemini \cite{GeminiTeamRST24} and Phi \cite{AbdinJAA24} in early experiments.
We provide complementary analysis for the discarded models in \Cref{lab:scenarios}.

\subsection{Mistake identification protocol (MIP)}
\label{sec:MIP}
We consider two prompting strategies to identify possible mistakes in a summary: direct and CoT prompting.
In \textbf{Direct prompting} \cite{TyenMCC24}, given the predicted summary and the meeting transcript, when required (see \Cref{tab:error_definition}), the model outputs 'Yes' or 'No' for each error to indicate its existence. 
For \textbf{CoT prompting} \cite{WeiWSB23b}, we extend direct prompting by having the model explain why a passage is erroneous following the 'let's think step by step' approach, allowing for detailed analysis of the model's understanding.

As GPT4 is not specifically trained to identify errors, we enrich the mistake identification prompt with few-shot examples of erroneous summaries (non-overlapping with our test set).
The mistake identification prompt consists of four parts: the model role and error definition for context, two few-shot examples of the error type, an optional request for the CoT prompting, and the primary task of reporting the error's existence.
We include more details on the exact prompt in \Cref{sec:prompts}.

We consider two setups to explore the MIP: a \textbf{single-instance} of a single GPT4 asked to detect all error types at once \cite{ZhangLZ23a} and a \textbf{multi-instance} architecture \cite{MousaviGRG23} using one GPT4 instance for each error type.

\subsection{Mistake identification discussion}
\label{sec:identification_discussion}
While both setups achieve high accuracy scores, the single-instance setup struggles to consistently beat an always true baseline on the whole QMSum Mistake dataset.
Overall, this aligns with the hypothesis behind current LLM-based automatic metrics that leverage similar models to assess text characteristics such as fluency, readability, or clarity \cite{LiXSX24}.

%\paragraph{How does the choice of mistake identification protocol affect the accuracy of pointing out errors?}
\paragraph{Impact of mistake identification protocol on accuracy of error detection.}
\input{tables/04_error_identification/breakdown_performance_feedback_identification}
Comparing results across the four MIP variants (\Cref{tab:subtableB}), we find that accuracy in detecting mistakes increases significantly on all error types when using a multi-instance setup compared to the single-instance approach.
While the difference between single and multi-instance is comparably small ($\sim$ 7\%) for both omission error types (T-OM, P-OM), the accuracy can deviate by up to $\sim$29.5\% in the case of HAL.
% S (d) - 68.4
% S (CoT) - 68.0
% M (d) - 81.9 | 80.7
% M (CoT) - 87.8 | 86.6
\Cref{fig:RQ1_mean_accuracy} shows that the average accuracy across all error types reveals a gain of at least 13.5\% when using multiple LLM instances for detection, which aligns with recent works  \cite{HuangCMZ24,TyenMCC24}.
We observe the average false negative rates decrease by $\sim$27\% from single (CoT) (30.0\%, worst overall) to multi (CoT) (3.4\%, best overall).
We hypothesize that the weaker single-model performance may stem from the extended content and its additional tasks, which must be handled by a single model compared to the multi-instance setting.
As a result, the single-instance approach is unable to process the long dependencies, which limits contextualization and comprehension \cite{LeeKHU21}.
In addition, while the multi-instance setup benefits from the CoT prompting, the single-model one is negatively affected with gains in false negative error rate.
The CoT explanations showing inconsistency in assessing requested error types due to misunderstanding of the definition supports these observations. % 

\input{figures/callables/RQ1_MeanAccuracyAcrossMistakes}

Considering the multi-instance approach as better suited for mistake identification, we conclude that the CoT prompting is beneficial to improve accuracy even further close to 90\%.
Note that the CoT explanation might contain wrong statements.
At the same time, the resulting error identification is correction, which has also been observed in tasks such as sorting and logical \cite{TyenMCC24}.

The nearly constant average false positive rate (between 12.4\% and 15.4\%) across all MIPs (\Cref{fig:RQ1_mean_accuracy}) suggests a model tendency to point out non-existing errors, which we interpret as oversensitivity to error types.
Analyzing the accuracy change between the whole dataset (\Cref{tab:subtableB}) and the erroneous subset (\Cref{tab:subtableA}) we find that GPT tends to falsely flag T-OM, P-OM, STR, HAL, and IRR errors. %...
We derive from this observation that the model expects a content-richer summary, seeing additional content as relevant.
% We expect that using the pointed-out additional content may lead to a more informative summary than the error-free human summaries when using this feedback for refinement in \Cref{sec:refinement}.
% Conclusion on findings
Our results suggest that the multi-instance setup with CoT prompting provides the most reliable mistake identification, which we use for the following experiments.

%\paragraph{What are the difficulties in identifying errors?}
\paragraph{Difficulties in identifying errors.}
%%%%%%
% Example for each division in appendix !!
%%%%%%
Based on the best MIP's accuracy, we categorize errors into three groups: \textbf{reliable} ($\geq$ 90.0\%: COR, REP, T-OM), \textbf{good} ($\geq$ 85.0\%: INC, LAN, STR), and \textbf{hard} to detect (\textless85.0\%: P-OM, IRR, HAL).
Following, we discuss the difficulties related to each category by analyzing the models' CoT explanations to identify patterns and the possible reason they struggle\footnote{Due to the amount of data, the model responses consi-\space dered for this section will be shared upon acceptance.}.

Errors from the reliable group have descriptions close to what an LLM without access to our definitions would generate when prompted to define the error.
The rare accuracy decreases are related to oversensitivity cases, e.g., assigning a T-OM error when expecting more details, indicating that the model may apply error detection rules too strictly.
False identification of COR errors typically occurs when conversations become less structured, and multiple participants mention similar information, as in the samples derived from the AMI dataset.

For errors from the good group, the main issue is the model's tendency to fail to properly contextualize error definitions and apply them too strictly compared to human annotators.
For example, a summary's linearity may be counted as a STR error, as the summary does not preserve the identical structure.
False detection of LAN errors includes marking domain-typical terms (e.g., grad student for graduate student in ICSI) as mistakes and orients on the transcript's language level, rendering fractured and brainstorming-like content (e.g., conversation from the ICSI) difficult.
% A similar issue with textual understanding arises for the INC error, where instances that might be seen as slightly incoherent are marked.

Errors from the hard group are challenging mainly due to the model's difficulty in understanding the error type.
In the context of HAL the model occasionally looks for closely related errors (e.g., T-OM, COR), leading to wrong detection.
% This happens comparable frequently for samples from the ICSI sub-dataset.
GPT4 struggles with P-OM and IRR due to the inherent subjectivity, which we also observe in the slightly lower inter-annotator agreement scores during the QMSum Mistake annotation (\Cref{tab:krippendorffs_alpha_human_annotation}).
We conclude that GPT4 applies error detection definitions slightly too strictly and mistakes related to subjectivity are influenced by the model's heuristic.

%% file: tables/04_error_identification/breakdown_performance_feedback_identification.tex
\begin{table*}[ht]
    \centering
    \small
    \begin{tabular}{cc}
        \begin{minipage}{0.48\textwidth}
            \centering
            \begin{tabular}{lccccc}
                \toprule
                & \multicolumn{2}{c}{\textbf{single-instance}} & \multicolumn{2}{c}{\textbf{multi-instance}} & \textbf{always true}\\
                \textbf{Error} & direct & CoT & direct & CoT & \\
                \midrule
                P-OM    & 75.0 & 82.2   &  82.5 & \textbf{84.5}  & 79.0\\
                T-OM    & 78.5 & 81.5   & 87.0  & \textbf{90.0}  & 81.0\\
                REP     & 73.0 & 72.0   & 92.0  & \textbf{95.5}  & 48.5\\
                INC     & 73.0 & 66.5   & 83.0  & \textbf{89.5}  & 39.0\\
                COR     & 76.0 & 63.0   & 85.0  & \textbf{91.5}  & 19.0\\
                HAL     & 42.5 & 59.0   & \textbf{73.5}  & 72.0  & 62.0\\
                LAN     & 61.5 & 68.5   & 77.5  & \textbf{88.5}  & 43.0\\
                STR     & 71.0 & 62.5   & 69.5  & \textbf{87.0}  & 47.0\\
                IRR     & 60.5 & 59.0   & 76.5  & \textbf{81.0}  & 51.0\\
                \bottomrule
            \end{tabular}
            \subcaption{Results on the whole QMSum Mistake dataset.}
            \label{tab:subtableB}
        \end{minipage} &
        \begin{minipage}{0.48\textwidth}
            \centering
            \begin{tabular}{lccccc}
                \toprule
                & \multicolumn{2}{c}{\textbf{single-instance}} & \multicolumn{2}{c}{\textbf{multi-instance}} & \textbf{always true}\\
                \textbf{Error} & Direct & CoT & Direct & CoT & \\
                \midrule
                P-OM    & 86.4  & 89.9  & 93.5 & \textbf{94.1} & 93.5\\
                T-OM    & 87.0  & 87.6  & \textbf{94.7} & 94.1 & 95.9\\
                REP     & 68.0  & 66.9  & 90.5 & \textbf{94.7} & 57.4\\
                INC     & 68.0  & 60.4  & 79.9 & \textbf{88.2} & 46.2\\
                COR     & 71.6  & 61.5  & 82.8 & \textbf{89.9} & 22.5\\
                HAL     & 50.3  & 60.4  & \textbf{75.7} & 75.1 & 73.3\\
                LAN     & 61.5  & 63.9  & 75.7 & \textbf{82.2} & 50.9\\
                STR     & 66.9  & 60.9  & 67.5 & \textbf{89.9} & 55.6\\
                IRR     & 55.6  & 60.4  & 76.9 & \textbf{82.8} & 60.4\\
                \bottomrule
            \end{tabular}
            \subcaption{Results on the erroneous samples of QMSum Mistake.}
            \label{tab:subtableA}
        \end{minipage}
    \end{tabular}
    \caption{Mistake identification accuracy of GPT4 for all MIP variants. Always True baseline provided for reference.
    Best values are \textbf{bold}.}
    \label{tab:breakdown_feedback_identification}
\end{table*}

%% file: figures/callables/RQ1_MeanAccuracyAcrossMistakes.tex
\begin{figure}[t]
  \includegraphics[width=1.0\linewidth]{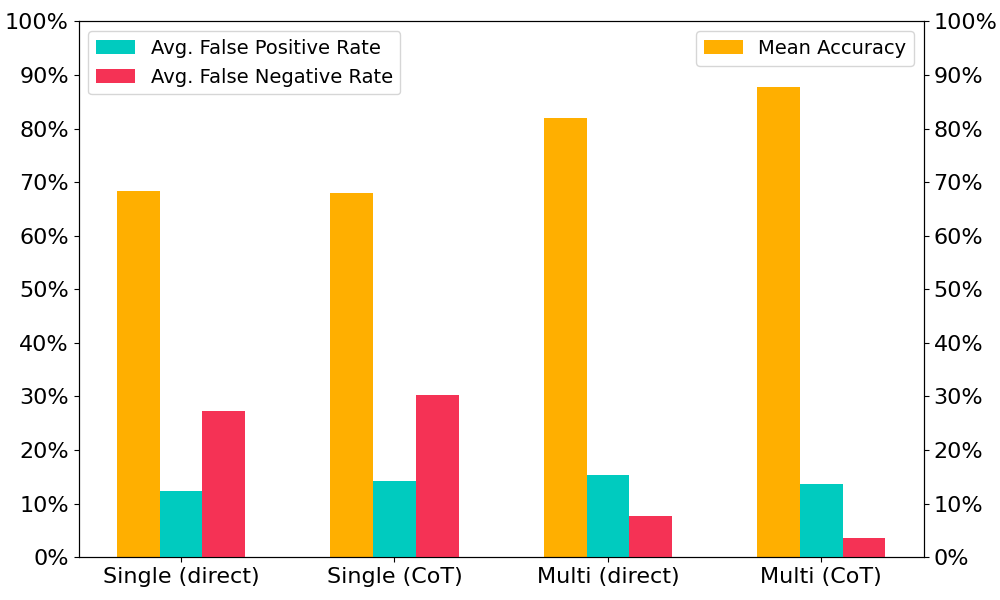}
  \caption {Average mistake identification accuracy, false positive and false negative rates for each MIP variant. For the accuracy, higher score is better. For the false positive/negative rate, lower is better.}
   \label{fig:RQ1_mean_accuracy}
\end{figure}

%% file: text/05_refinement.tex
Building on the finding that an LLM can identify typical meeting summarization errors (\Cref{sec:identification_discussion}), we analyze how the quality of original predicted summaries changes when an LLM refines them based on identified mistakes.
Our multi-model refinement approach mimics a four-stage human review process to form a refinement protocol (\Cref{fig:overview_setup}): (1) locating errors using the best-performing MIP, (2) generating feedback on identified errors (\textbf{feedback protocol}), (3) structuring feedback (\textbf{transfer protocol}), and (4) refinement.
Following, we explore the setup of the feedback and transfer protocols to derive a refinement protocol for meeting summarization.

\subsection{Feedback protocol (FP)}
\label{sec:FP}
Feedback on an error can range from pointing out its existence, similar to someone highlighting a text passage and leaving a short comment, to in-depth explanations of what is wrong with the marked passage and rewrite suggestions.
Following this analogy, our feedback protocol consists of an \textbf{essential} and an \textbf{additional detail} part.
The essential part includes minimal feedback on the existence of an error type and a short explanation about why and where it was detected, but may not mention all error instances.
The additional detail part considers three optional information sources: \textbf{CoT explanation} \cite{WeiWSB23b}, \textbf{correction suggestion} \cite{ZhangLZ23a}, and the original \textbf{transcript}.
CoT explanation, the output of MIP's CoT prompting (\Cref{sec:MIP}), contains all observed error instances and details on why they are considered errors.
It helps the refinement model derive a rewriting plan through detailed, structured information but may lead to confusion if the reasoning is wrong \cite{TyenMCC24}.
Correction suggestions provide examples of how to correct the error, either as tips or precise rewrites that can be directly applied.
The transcript provides all available information in its original form, allowing it to decide whether to accept or reject the feedback and how to integrate it.
The three optional information sources can be combined, determining how much information is required and if feedback without a transcript is as informative as adding the transcript for lookup.
% See \Cref{sec:prompts} for details on the protocol variants.

\input{tables/04_error_identification/refinement_ranking}

\subsection{Transfer protocol (TP)}
\label{sec:TP}

We consider two approaches for structuring feedback for the refinement model: \textbf{direct feedback} \cite{MousaviGRG23} and \textbf{consolidation} \cite{ZhangLZ23a}.
Direct feedback transfers derived feedback without additional processing, stating whether an error type is observed or not.
In the case of CoT explanation, it informs the model step-by-step which sentences are erroneous or error-free, why they are correct or incorrect, and what should be changed (or kept) to have a correct sentence.
Consolidation considers only identified errors and generates an editing plan using an intermediate LLM, extracting what information to add, remove, or alter from the feedback protocol.
The consolidation protocol does not affect an appended transcript.

% We should have an example on this how it looks before and after the consolidation in the appendix maybe?
% Intermediate LLM Output: "Add: Specific project deadlines. Remove: Redundant mentions of 'the budget'. Alter: Rephrase 'the budget, the budget allocation' to 'the budget allocation'."

\subsection{Experimental setup}
We refine the erroneous summaries from QMSum Mistake using each refinement protocol variant with the multi-instance CoT-prompted MIP.
GPT4 is used as the backbone model for the refiner and optional intermediate LLM to consolidate feedback, with other model families explored in \Cref{lab:scenarios}.
We focus the following experiment on evaluating how summary quality changes based on feedback and show a setup for a meeting summarization refinement protocol.
We consider a one-shot improvement here and provide insights on multi-round improvement in \Cref{sec:multiple_rounds}.
To help understand and categorize the quality changes, we report metric results for the original erroneous summaries \textit{(ORIG)}, error-free QMSum gold summaries \textit{(GOLD)}, summaries generated by one GPT4 \textit{(GPT-S)}, and summaries refined by one GPT4 \textit{(GPT-R)}\footnote{'Refine this summary by considering the transcript'.} as references in \Cref{tab:refinement_ranking}.
%Despite allowing the generation of 4k tokens as feedback, the average feedback length is 2.2k tokens.

\subsection{Evaluation approach}
\label{sec:refinement-evaluation}

ROUGE \cite{Lin04b} and BERTScore \cite{ZhangKWW20b}, established metrics for meeting summary evaluation \cite{KirsteinWGR24}, yield scores too similar for interpretation across protocol variants (see \Cref{tab:automatic_metric_scores_refinement}).
As human evaluation on all generated refined summaries (total $\sim$3.4k) is infeasible, we use the LLM-based metric AUTOCALIBRATE \cite{LiuYHZ23} to report Likert scores on relevance (REL), informativeness (INF), conciseness (CON), and coherence (COH).
Since this metric is not developed for meeting summarization, we assess alignment with human judgment by having six annotators rate a subset of 200 summaries according to AUTOCALIBRATE prompts (inter-annotator agreement (Krippendorff's alpha): REL: 0.775, INF: 0.798, CON: 0.833, COH: 0.803). 
As the LLM-based evaluation aligns sufficiently with annotator labels (accuracy: 89.1\%), we use AUTOCALIBRATE as our main quality proxy.
Nevertheless, we manually check every fourth score tuple and model reasoning to confirm alignment with the evaluation task and human judgment.
In case of misalignment, three annotators would instead rate the summary.
As AUTOCALIBRATE only assesses specific characteristics and does not consider omission, hallucination, or repetition, we also set up a GPT4-powered ranking system, motivated by typical human annotation rankings, based on observable errors from \Cref{sec:errors} (see \Cref{sec:prompts} for prompt details).
We follow the approach used before to ensure reliability and alignment with human annotations (inter-annotator agreement: 0.784 Krippendorff's alpha, GPT4 acc.: 92.1\%).

\subsection{Summary refinement discussion}
\paragraph{Influence of feedback and transfer protocols on quality.}

\Cref{tab:refinement_ranking} shows the overall ranking and Likert scores of each refinement protocol variant.
ORIG summaries are consistently ranked lowest, indicating that refinement positively influences quality, as observed in the assigned Likert scores.

Having only the essential part in the FP leads to minor improvements in ranking and Likert scores for both TPs compared to the ORIG summary, but falling behind the scores of most protocol variants using additional information.
This indicates that pointing out errors on a high level already leads to quality improvement.
The result is expected, as the minimalistic explanation may not contain every error instance, precise reasoning, or all information to resolve specific errors such as omission.
Comparing the essential parts scores of both TPs reveals that the Likert scores and rankings differ notably between the two with the scores using consolidated TP being $\sim$ 0.7 points less.
We derive from this observation that the provided feedback influences scores and leads to quality changes.

For the direct TP, CoT explanation and correction are ranked higher (avg. ranks $\sim$3.75) than the GPT-S summaries (avg. rank 4.84) and are close to GOLD summaries (avg. rank 4.04).
CoT explanation and correction-based refinement outperform transcript-based refinements in overall ranking (avg. rank 4.68 to 5.10) but fall behind in Likert scores, which appears counter-intuitive.
The ranking LLM's reasoning reveals that transcript-based refinement contains repetitions, fails to separate topics, and lacks details, leading to an overall worse rating compared to CoT and correction.
As the longer prompt when providing transcripts (avg. 20k tokens with transcript, 4k tokens without transcript) is still only a sixth of GPT4's context size, we hypothesize that the additional task of cross-checking errors with the transcript may confuse the model due to content repetition and noise in the form of unnecessary details.
CoT explanation and correction appear as a lean alternative containing relevant information for quality improvement.

CoT explanation and correction (avg. ranks $\sim$3.75) both outperform the combined use of the two (avg. ranking of 5.1 with and 4.11 without transcript).
The analysis of the ranking model's explanation shows that the repetition of content in CoT and correction can lead to multiple occurrences of the same information, while contradicting content may lead to the inclusion of wrong information (see an example in \Cref{fig:confusion_CoT_Corr}).

For the consolidation TP, FPs without transcripts barely improve summary quality (avg. ranks range from 5.61 to 6.40).
Transcript-using variants perform similarly to their direct TP counterparts but with rankings and scores closer together.
Further, their scores are close to the GPT-R results.
This indicates that the consolidated feedback has less influence on refinement than the direct feedback and that the refinement model relies more on the transcript to rewrite the summary.
The refiner model's reasoning reveals that the refinement approaches with consolidation TP and without transcript access often omit details and lack conciseness, which is also observable in the Likert scores (e.g., CON up to 0.47 points down).
We conclude that the consolidated approach, effective in short news summarization \cite{ZhangLZ23a}, does not perform well for meeting summarization, likely because the format compresses information about individual errors too much, making it hard for the refinement model to interpret when the total number of errors is large.
This can happen especially with long meetings (16k tokens input text) compared to news summarization with input texts of 200 tokens.
We leave to future work the exploration of a similar planning setup that applies the described consolidation structure to each error-related feedback block rather than to the whole feedback

We conclude that the feedback from the MIP containing CoT explanations already provides a strong foundation for improving ORIG summaries and bringing their quality close to that of a human summary.
Correction suggestion is a promising alternative for CoT explanation as FP with comparable ratings on quality, allowing for further research to identify when to use which FP.

%% file: tables/04_error_identification/refinement_ranking.tex
\begin{table*}
  \centering
  \small
  \begin{tabular}{llccccc}
    \toprule
    \textbf{TP} & \textbf{FP} & \textbf{Overall} & \textbf{REL} & \textbf{INF} & \textbf{CON} & \textbf{COH} \\
    & & (Ranking $\downarrow$) & (Likert $\uparrow$) & (Likert $\uparrow$) & (Likert $\uparrow$) & (Likert $\uparrow$) \\
    \midrule
    \multirow{8}{*}{direct} & essential only  & 5.44 & 3.08 & 2.99 & 3.29 & 3.14  \\
    & CoT &  \underline{\textbf{3.75}} & 3.10 & 3.14 & 3.46 & 3.20 \\
    & Cor &  3.79 & 3.04 & 2.83 & 3.57 & 3.23 \\
    & CoT+Cor  & 4.11 & 3.11 & 2.88 & 3.40 & 3.09 \\

    & Tra  & 4.68 & 3.12 & 2.93 & 3.65 & 3.37 \\
    & Tra + CoT & 4.74 & \textbf{3.14} & \underline{\textbf{3.36}} & 3.67 & \textbf{3.56} \\
    & Tra + Cor & 4.93 & 3.10 & 3.14 & \textbf{3.68} & 3.44 \\
    & CoT+Cor+Tra  & 5.10 & 3.05 & 3.05 & 3.43 & 3.18\\
    \midrule
    \multirow{8}{*}{consolidated} & essential only  & 6.10 &  2.53 & 2.27 & 2.58 & 2.36\\
    & CoT  & 5.61 & 2.69 & 2.62 & 2.99 & 2.70\\
    & Cor  & 6.07 & 2.96 & 2.85 & 3.22 & 2.98\\
    & CoT+Cor  & 6.40 & 2.93 & 2.92 & 3.34 & 3.03\\

    & Tra  & \textbf{4.86} & 3.08 & 3.12 & 3.50 & 3.33\\
    & Tra + CoT & 4.89 & 3.04 & 3.05 & 3.49 & 3.22 \\
    & Tra + Cor & 4.88 & 3.11 & \textbf{3.29} & 3.60 & \underline{\textbf{3.59}} \\
    & CoT+Cor+Tra  & 4.92 & \underline{\textbf{3.21}} & 3.18 & \underline{\textbf{3.70}} & 3.46\\
    \midrule
    \multirow{3}{*}{} & GOLD & 4.04  & 3.08 & 3.05 & 3.53 & 3.21\\
    & ORIG & 6.75 & 2.28 & 2.15 & 2.41 & 2.22\\
    & GPT-S & 4.84 & 3.00 & 3.00 & 3.40 & 3.10\\
    & GPT-R & 4.82 & 3.09 & 3.09 & 3.72 & 3.44\\
    % \midrule
    % \textbf{Descriptive Statistics} & \textbf{Mean (Likert)} & - & - & 3.00 & 2.80 & 3.20 & 3.10 \\
    % & \textbf{Std (Likert)} & - & - & 0.27 & 0.35 & 0.35 & 0.35 \\
    \bottomrule
  \end{tabular}
  \caption{Quality reporting of refined summaries for all Transcript Protocols (TP) and Feedback Protocols (FP) combinations (CoT = CoT explanation, Cor = correction, Tra = Transcript). Ranking is the average ranking across all samples. Lower ranking scores indicate higher preference (1 (always preferred) to 20 (always disliked)). REL, INF, CON, COH are the AUTOCALIBRATE Likert scores on relevance, informativeness, conciseness, and coherence using a 5-step Likert scale (1 (worst) to 5 (best)). Best scores per TP are \textbf{bold}, best scores overall are \underline{underlined}.}
  \label{tab:refinement_ranking}
\end{table*}

%% file: text/A_HumanAnnotation.tex
We adapt proven methodologies for a thorough human annotation \cite{ZhangLYF23}.
Six graduate students \footnote{The origin of the funds and annotators will be disclosed later to avoid the risk to give the authors identity.} aged 22 to 28, from diverse academic backgrounds (e.g., computer science, psychology, communication science), proficient in English and familiar with meeting summarization, participate as annotators.
Each annotates 2-4 subsets of QMSum Mistake (35 samples per subset, each containing transcripts, gold summaries, and model-generated summaries), with at least three annotators per sample.
The annotators identify errors by answering yes/no questions (e.g., "Does the summary contain repetition %of content which does not add to the contextualization or understanding
?") and provide reasoning for observed mistakes for quality assessment.
To ensure reliable and consistent annotation, we implement several measures.
Inter-annotator agreement is assessed using Krippendorff’s alpha \cite{Krippendorff70}, achieving an average of 0.764 (see \Cref{tab:krippendorffs_alpha_human_annotation}), indicating a moderate to strong agreement on the assessment.
A training course is run before the annotation task to train annotators, refine guidelines, and identify new error types.
Annotators practice on QMSum summaries produced by the LED model, which are not used for the final QMSum Mistake dataset.
During the actual annotation task, we add gold summaries that the annotators should be able to evaluate correctly.
Otherwise, their annotation would be rejected, and their understanding of the task would be discussed.
Regular review meetings are held to maintain consistency in quality, align understanding, and conduct ongoing quality control.
An expert annotator was available to discuss complex issues during the annotation process.
\input{tables/03_dataset/inter-annotator_agreement}

%% file: tables/03_dataset/inter-annotator_agreement.tex
\begin{table}
  \centering
  \small
  \begin{tabular}{lc}
    \toprule
    \textbf{Assessed Characteristic} & \textbf{Krippendorff's $\alpha$} \\
    \midrule
    Omission (partial)   & 0.787  \\
    Omission (total) & 0.834  \\
    Repetition & 0.889  \\
    Incoherence     & 0.764   \\
    Coreference     & 0.719   \\
    Hallucination     & 0.764  \\
%    Hallucination (intrinsic)     & 0.764  \\
%    Hallucination (extrinsic)     & 0.757  \\
    Language   & 0.748  \\
    Structure  & 0.795   \\
    Irrelevance & 0.719 \\
    \bottomrule
  \end{tabular}
  \caption{Inter-rater reliability for the human annotations, measured by Krippendorff's alpha. Scores $\geq$ 0.667 mean moderate agreement, scores $\geq$ 0.8 mean strong agreement.}
  \label{tab:krippendorffs_alpha_human_annotation}
\end{table}

%% file: text/B_AdditionalModelFamilies.tex
\label{lab:scenarios}
In this section, we task models from the Phi and Gemini families on the mistake identification and refinement tasks.
Particularly, we consider Gemini Flash (Gemini) and the 3.4B parameter Phi-3 mini 128k (Phi).
We chose these models because their context size is large enough to fit a meeting transcript without requiring major architecture adaptation and because they are available.
We further opt for smaller model versions compared to GPT4 to analyze the performance differences.
We perform the experiments on 25\% of the erroneous QMSum Mistake samples to derive initial trends.

% \begin{table}
%     \centering
%     \small
%     \centering
  
%     \begin{tabular}{lcccc}
%         \toprule
%         \textbf{Error Type} & \textbf{GPT-4} & \textbf{Phi} & \textbf{Gemini} & \textbf{Claude}\\
%         \midrule
%         OMI (partial)   & 0.864 & 0.864 & 0.864 & 0.864 \\
%         OMI (total) & 0.870 & 0.864 & 0.864 & 0.864 \\
%         REP & 0.680 & 0.864 & 0.864 & 0.864\\
%         INC     & 0.680 & 0.864 & 0.864 & 0.864\\
%         COR     & 0.716 & 0.864 & 0.864 & 0.864\\
%         HAL (intrinsic)     & \textbf{0.580} & 0.864 & 0.864 & 0.864\\
%         LIN  & 0.615 & 0.864 & 0.864 & 0.864\\
%         STR    & 0.669 & 0.864 & 0.864 & 0.864\\
%         IRR & 0.556 & 0.864 & 0.864 & 0.864\\
%         \bottomrule
%     \end{tabular}

%   \caption{Mistake finding accuracy for all models. Error Types as in previous table}
%   \label{tab:feedback_different_models}
% \end{table}

% \input{figures/callables/RQ2_template_result} % --> alternative to table

\subsection{Mistake Identification with smaller models}
\begin{table}[ht]
    \centering
    \small
      \begin{tabular}{lccccc}
        \toprule
        \textbf{Error} & \textbf{Gemini} & \textbf{Phi} & \textbf{GPT4}\\
        \midrule
        P-OM    & 87.5  & 87.5 & 87.5\\
        T-OM    & 75.0  & 75.0 & 92.5\\
        REP     & 35.0  & 32.5 & 90.0\\
        INC     & 62.5  & 32.5 & 95.0\\
        COR     & 15.0  & 7.5 & 92.5\\
        HAL     & 57.5  & 57.5 & 57.5\\
        LAN     & 35.0  & 35.0 & 72.5\\
        STR     & 37.5  & 20.0 & 92.5\\
        IRR     & 60.0  & 60.0 & 77.5\\
        \bottomrule
      \end{tabular}
      \caption{Mistake finding accuracy of Gemini, Phi, GPT4 on a subset of QMSum Mistake.}
      \label{tab:feedback_different_models}
\end{table}

\Cref{tab:feedback_different_models} shows the accuracies of these models in terms of identifying errors, all using the best MIP protocol identified in \Cref{sec:MistakeIdentification}, containing multiple model instances and CoT prompting.
As expected, Gemini and Phi show weaker accuracy, which can mostly be attributed to their smaller model sizes.
Notably, Phi struggles to report errors in the prompted output format, similar to how GPT4 struggles in the single-instance setup, while Gemini is closer in its answer pattern to what we observed for GPT4 in the single-instance setup.
Phi and Gemini also show an oversensitivity to errors as we hypothesize for GPT4 (\Cref{sec:identification_discussion}).
This oversensitivity is more pronounced for the smaller Phi model than for Gemini.
This oversensitivity leads to a match in accuracy for P-OM and HAL, as all models reported here an always-true result.
Considering the models' reasoning for the scores, we observe further support for this hypothesis.
For example, Gemini reports the mention of participants' names as an unnecessary repetition.
We conclude that even though these models have a similar (Phi) or larger (Gemini) context size compared to GPT4, the significantly fewer parameters hurt the task understanding and contextualization.
Further, the oversensitivity appears to be linked to a model's understanding capabilities, which in the considered case is connected to the model size.

\subsection{Refinement Performance with Smaller Models}
\Cref{tab:qualit_changes_phi} reports the quality of one-round refined summaries using Phi and GPT4 on the subset of QMSum Mistake.
Note that GEMINI is not reported here as the model consistently did not provide any refinements.
Both models were prompted with the best-performing refinement protocol, i.e., multiple instances of CoT were prompted for mistake identification, CoT explanation was used as feedback, and direct feedback was used as a transfer protocol.
We follow the evaluation approach in \Cref{sec:refinement-evaluation}.
We observe that even though Phi does not reliably detect errors, the exhaustive pointing out of possible error cases and the refinement step help to improve the quality, considering the Likert scores by 0.4 to 0.8 points.
However, it is noteworthy that Phi sometimes struggles with refining a summary and instead details the given feedback.
We therefore conclude, that Phi is capable of refining a summary given a list of observed errors and reasoning for the observation, but the smallest model struggles with the task understanding.
Hence, with adaptions such as few-shot examples or by using Phi-3 small, Phi may be a cheap alternative to GPT4 for summary refinement.

\begin{table}
\centering
\small
\begin{tabular}{lccccc}
\toprule
 & \textbf{OVR $\downarrow$} & \textbf{REL $\uparrow$} & \textbf{INF $\uparrow$} & \textbf{CON $\uparrow$} & \textbf{COH $\uparrow$}\\
% \textbf{TP} & \multirow{6}{}{-} & \multirow{6}{}{CoT} & \multirow{6}{}{Cor} & \multirow{6}{}{CoT+Cor} &&\\
\midrule
GPT4 & 1.24 & 3.05 & 3.07 & 3.21 & 2.98 \\
Phi & 1.84  & 2.78 & 2.98 & 2.93 & 3.04 \\
\midrule
GOLD & 1.43 & 3.08 & 3.05 & 3.53 & 3.21 \\
ORIG & 2.77 & 2.28 & 2.15 & 2.41 & 2.22 \\
\bottomrule
\end{tabular}
\caption{Ranking and scoring of Phi and GPT4 according to their quality. OVR is the overall ranking, with lower scores indicating a more preferred summary. REL, INF, CON, and COH are relevant, informativeness, conciseness, and coherence. The scoring uses a 5-step Likert scale, with 1 being the worst and 5 best.}
\label{tab:qualit_changes_phi}
\end{table}

\subsection{Multiple rounds}
\begin{figure*}
    \centering
    \includegraphics[width=1.0\linewidth]{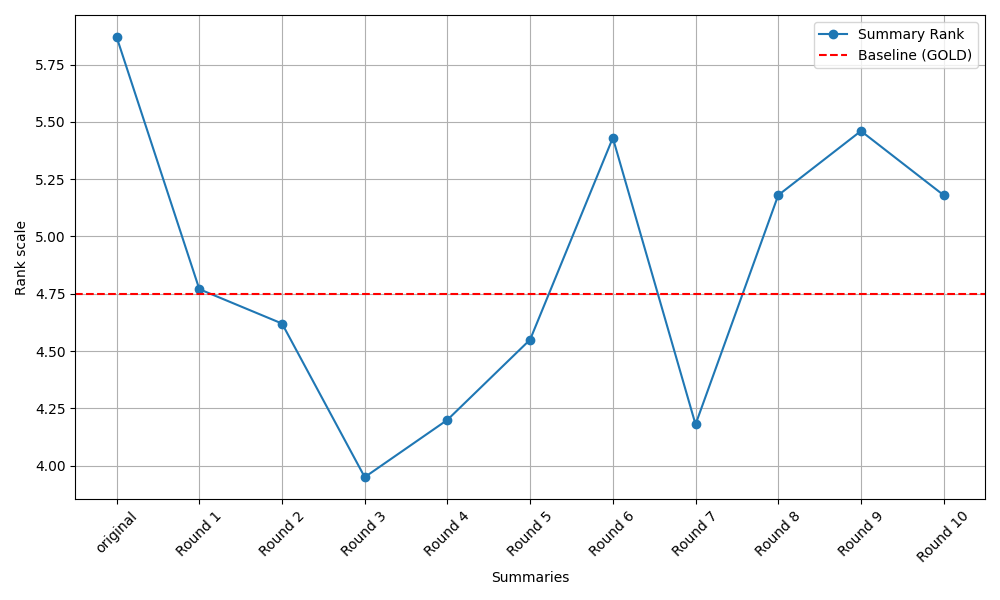}
    \caption{Ranking of multiple summaries refined for up to 10 rounds. The red dotted line indicates the ranking of the GOLD summaries.}
    \label{fig:multi-round}
\end{figure*}

So far we have explored the application of the refinement concept in a single round, with one pass of the mistake identification and summary refinement.
Following, we explore how the refinement quality changes when GPT4 can reconsider the generated summary for 10 rounds.
We keep the best-performing setup (multi-instance with CoT prompting for MIP, CoT explanation FP, direct feedback TP) and use the small subset of QMSum Mistake.
We report the ranking of the different summaries in \Cref{fig:multi-round}, observing that while the one-round performance is strong enough to improve a given summary to a quality level comparable to a human summary, it can be further improved.
From the ranking model's reasoning, we observe that this improvement mainly involves reducing remaining omission errors and fitting the summary better to the comprehensiveness GPT4 asks for.
Notably, we observe instances of strong degradation, e.g., in 6 which follows a previous trend of reduced quality.
We derive from this that while there may be more potential to further improve summaries by applying the refinement protocol multiple times, it may quickly saturate, and unwanted errors are induced.
From the ranking model's explanation, we observe that this correlates with an increase in repetition and hallucination.
We conclude that multiple rounds of refinement can potentially further improve summaries, but this requires dedicated research.

\label{sec:multiple_rounds}

%% file: text/C_QMSum_Mistake_Examples.tex
\begin{table}
    \centering
    \small
    \begin{tabular}{llcccc}
    \toprule
    TP & FP      & BS & R-1 & R-2 & RLS \\
    \midrule
    dir & essential     & 16.20 & 33.73 & 07.46 & 20.53 \\
    dir & CoT           & 16.16 & 33.89 & 07.57 & 20.41 \\
    dir & Cor           & 16.19 & 33.89 & 07.52 & 20.39 \\
    dir & CoT+Cor       & 16.35 & 33.90 & 07.56 & 20.58 \\
    dir & Tra           & 15.28 & 33.89 & 07.82 & 20.99 \\
    dir & Tra+CoT+Cor   & 15.12 & 33.78 & 07.94 & 21.31\\
    \midrule
    con & essential     & 14.27 & 29.79 & 05.58 & 18.26 \\
    con & CoT           & 14.28 & 29.36 & 05.43 & 18.12 \\
    con & Cor           & 15.11 & 29.64 & 05.55 & 18.37 \\
    con & CoT+Cor       & 15.15 & 29.71 & 05.71 & 18.13 \\
    con & Tra           & 14.96 & 29.90 & 05.55 & 18.55 \\
    con & Tra+CoT+Cor   & 14.98 & 30.07 & 05.76 & 18.47 \\
    \bottomrule
    \end{tabular}
    \caption{Score of the established evaluation metrics BERTScore (BS) and ROUGE (R-1 = ROUGE 1, R2 = ROUGE 2, RLS = ROUGE LSum).}
    \label{tab:automatic_metric_scores_refinement}
\end{table}

We show one examples of QMSum Mistake for each used language model in \Cref{tab:qmsum_mistake_examples}.

\begin{table*}[]
    \centering
    \small
    \begin{tabular}{lp{1.8cm}p{8.5cm}l}
    \toprule
    Architecture & Language Model  & Summary & Errors\\
    \midrule
    enc-dec 
                    & LED       & The group discussed the components design of the remote control. Firstly, they discussed the shape of the buttons, the size of the screen, the colour of the rubber case, and the material. Then, they talked about the material design. After that, the group discussed how the remote should look like a fruit and vegetable. Finally, the team discussed the design of a touch-based graphical interface system. The team decided to use the wheely material instead of a rubber case for the LCD screen. Then the group talked about how the material should be used for the remote, and how it should be easy to switch to. & \makecell[{{p{1.7cm}}}]{REP, LAN \\ INC, P-OM \\ T-OM, HAL} \\
    &&&\\
                    & DialogLED & This was the kick-off meeting for the new remote control project. Project Manager started the meeting by introducing the team members to each other. Industrial Designer introduced the group as the project manager. Marketing and User Interface gave a brief about the project and the design of the remote control. Then Industrial Designer gave the brief about their favourite characteristics of their favourite animals. Then the group got to draw their favourite animal and sum up their favorite characteristics of it. Industrial designer gave the final product and the team got to choose the favourite animal of the group. &  \makecell[{{p{1.7cm}}}]{REP, LAN \\ INC, P-OM \\ T-OM, COR \\ HAL, STR} \\
    &&&\\
                    & PEGASUS-X & Project Manager introduced some new project requirements that came in, and then Industrial Designer discussed the working design of the remote control. User Interface and Industrial Designer talked about the buttons and what it does and that sort of thing. Project Manager discussed the target age group and how it's gonna appeal to their target group. Marketing and User Interface discussed the functions of the circle disc, the touch pad, and the volume and channel changers. Industrial Designer said that it would be easy to implement and that it'd be compatible with D-V-D-type remotes. & \makecell[{{p{1.7cm}}}]{LAN, P-OM \\ T-OM} \\
    \midrule
    autoreg 
                    & GPT-3.5 & The conversation is about a team discussing the design of a new remote control for a television. The project manager mentions that the remote control should be original, trendy, and user-friendly. The team discusses different ideas for the remote control, including using speech recognition for navigation and having a light adaptation system. They also discuss the importance of optimizing the number of buttons and the size of the remote control. The industrial designer suggests gathering data from programs or content and navigating through them using the remote control. The team agrees to discuss technical points and gather more information in the next meeting. They also mention the need to assign roles and responsibilities for different aspects of the design. The conversation ends with the project manager suggesting a follow-up meeting in five minutes to discuss other aspects of the design. & \makecell[{{p{1.7cm}}}]{P-OM, T-OM\\ HAL, IRR} \\
    &&&\\
                    & Phi-3 mini & The group discussed the progress on digits, specifically the almost completion of the TI-digits project and the plan to train a recognizer with the data. They also talked about the possibility of having transcribers manually input the digits from a meeting recording, which would result in a more complete transcript. There was a debate on whether this task should be done by the transcribers or automated. Additionally, they touched on the quality of the current models and the potential for improvement, as well as the differences in pronunciation when reading digits compared to conversational speech. They considered using Switchboard data for further research and discussed the need for a more comprehensive annotation system that includes articulatory features. & \makecell[{{p{1.7cm}}}]{P-OM, T-OM \\ HAL} \\
    \bottomrule
    \end{tabular}
    \caption{Samples of the QMSum Mistake dataset, one for each used language model. In the architecture column, enc-dec means encoder-decoder and autoreg stands for autoregressive. The errors column presents the human-annotated errors for each summary.}
    \label{tab:qmsum_mistake_examples}
\end{table*}

%% file: text/D_Prompts.tex
\label{sec:prompts}

In the following, we present the prompts used to identify mistakes (\Cref{fig:MIP_prompt}), to consolidate feedback (\Cref{fig:consolidate_prompt}), and to refine a summary (\Cref{fig:refinement_prompt}).
\Cref{fig:few_shot_example_prompt} shows a few-shot example of P-OM.
\Cref{fig:llm-ranking} provides the prompt template for LLM-based ranking.

\begin{figure*}[t]
    \begin{AIbox}{Multi-Instance Protocol Prompt Template}
    \parbox[t]{\textwidth}{
        You are an experienced linguist and you will be given one summary for a meeting.
        Your task is to rate the summary based on the existence of the below-provided error type.
        Please make sure you read and understand these instructions carefully.
        Please keep this document open while reviewing, and refer to it as needed.
        Following is the error type(s) you should look for: \newline 
        """{error definition}""" \newline
        \newline
        Evaluation steps: \newline
        1. Read the transcript, if available, carefully and identify the main topic and key points. \newline
        2. Read the predicted summary and compare if it contains instances of the described error type. Note every instance you observe that is part of the error type. Only consider the error type and no other mistakes else. \newline
        3. Rate the summary based on the existence of the error type with yes when at least one instance of the error type is found or no if the summary does not exhibit the error type. (primary task). \newline
        4. You may be given secondary tasks, such as thinking step by step, explaining your decision, or pointing out the locations of each individual instance of the error type. These secondary tasks are designed to help you become more certain about your decision. \newline
        5. Provide your findings in the desired format, so that your final output is a report on the existence of the error type in the given summary. \newline
        Tip: Consider the whole input, i.e., the transcript and the predicted summary, provided in the user's prompt to make a good decision that humans will agree on. \newline
        Below are two examples demonstrating the different impact levels of the previously described error type. Please learn from these examples the concept and how the rating works. \newline
        Example 1: """{minor error example prompt}""" \newline
        Example 2: """{major error example prompt}"""\newline
        \newline
        Your secondary task: """{e.g., Let's think step by step and describe every step you consider which leads you to the result that an error occurs or not.}""" \newline
        Your primary task: """Please provide feedback on the existence of the error. Does this passage contain an error? Answer 'yes' or 'no'.""" \newline
        \newline
        You should now perform the error search on the following predicted summary: """{summary}""" \newline
        (optional) If required, you can use the original transcript for look up: """{transcript}""" \newline
        Please follow the following structure for your output and fill in the blanks: """{format}""" \newline
    }
    \end{AIbox}
    \caption{MIP prompt template in the format for multi-instance usage. In the single-instance setup, the definition and example blocks are repeated for every error type.}
    \label{fig:MIP_prompt}
\end{figure*}

\begin{figure*}[t]
    \begin{AIbox}{Feedback Consolidation Prompt Template}
    \parbox[t]{\textwidth}{
       You are a professional feedback summarizer, that provides a comprehensive, direct version of a feedback report.
       Your condensed version should be usable for someone to improve their previous summary effectively.
       So you are allowed to structure it in the most effective way to address the feedback. The refinement should be successful purely from your feedback and the previous summary so include all relevant details given in the report. \newline
       \newline
        Please consolidate the following feedback into a plan and provide usable feedback: """{positive feedback}""". \newline
        Use the output format 'Add: <Add the information of ...>  Remove: <Remove the information of ...>  Rephrase: <Rephrase the information of ...> Simplify: <Shorten the summary regarding ...> Keep: <Keep the summary unchanged at ...>'.
        Include all details from the feedback.
    }
    \end{AIbox}
    \caption{Prompt tehmplate used to consolidate a feedback for the consolidation TP. The model is tasked to extract from the exhaustive feedback what the refinement model should consider for editing.}
    \label{fig:consolidate_prompt}
\end{figure*}

\begin{figure*}[t]
    \begin{AIbox}{Summary Refinement Prompt Template}
    \parbox[t]{\textwidth}{
        You are an expert in refining and improving summaries.
        Your task is to improve the summaries of conversations based on a given feedback report.
        All the content to improve the original summary and make it the very best is provided in the review, as the reviewer provides all details.
        \newline
        Please improve this summary: """{summary}""" \newline
        considering this review: """{feedback}""" 
    }
    \end{AIbox}
    \caption{The summary-refining sub-prompt.}
    \label{fig:refinement_prompt}
\end{figure*}

\begin{figure*}[t]
    \begin{AIbox}{Partial Omission Few-Shot example}
    \parbox[t]{\textwidth}{
        Transcript: """"Good morning, everyone. Today, we need to address the proposed increase in the marketing budget. After analyzing current trends and performance, the proposal is to increase the marketing budget by 50\% in Europe. This increase will primarily fuel our new digital marketing campaign targeting Europe. We believe this strategic focus will significantly boost our sales, and we plan to reassess this move after the first quarter to evaluate its impact on our growth metrics.""" \newline
        Predicted Summary: """The committee agreed to increase the marketing budget to support new initiatives.""" \newline
        Explanation: """This example shows high severity partial omission because the summary fails to specify the significant increase percentage, the targeted geographical focus of the marketing campaign, and the strategic plan for reassessment. These omissions leave out critical details necessary for understanding the scope and strategic intent of the budget increase, which could lead to significant misalignment in expectations and preparations among team members.""" \newline
    }
    \end{AIbox}
    \caption{A few-shot example as it is shown to the mode in the MIP prompt \Cref{fig:MIP_prompt}. This few-shot examples counts a major P-OM example.}
    \label{fig:few_shot_example_prompt}
\end{figure*}

\begin{figure*}[t]
    \begin{AIbox}{LLM-based Ranking}
    \parbox[t]{\textwidth}{
        You are an expert in the field of summarizing meetings and are tasked with evaluating the quality of the following summaries.
        Rank the following summaries based on their quality, with 1 being the best summary and 8 being the worst summary. \newline
        \newline
        Summaries to rank: \newline
        Transcript: """{transcript}"""\newline
        Summary 1: """{summary 1}"""\newline
        ... \newline
        Summary n: """<{summary n}"""\newline
        \newline
        The criteria for ranking the summaries include:\newline
        1. The summary should not contain any content-wise redundant information, that does not aid the understanding or contextualization. \newline
        2. The summary should be coherent, maintain logical flow, relevance, and clarity within a sentence and across sentences. \newline
        3. The summary should use appropriate language with correct and grammatical use. Language should not be ambiguous. \newline
        4. The summary should not ommit relevant content. Neither should content be completely absent or relevant details be missing. \newline
        5. The summary should correctly reference statements and actions to the respective meeting participant. \newline
        6. The summary should not add hallucinated content. This includes the additional of new content not present in the transcript as well as changing details. \newline
        7. The summary should maintain the logical and temporal structure and not misplace topics or events. \newline
        8. The summary should not contain irrelevant information but focus on what is important. \newline
        When encountering issues with any of these criteria, assess the impact and rate accordingly. Omission and hallucinated content are more severe issues than the other. \newline
        \newline
        Your task is to rank the summaries based on the criteria provided.
        Remember to consider the quality of the summaries and how well they capture the key points of the original transcript.
        First provide an argumentation for your ranking. Therefore, use chain-of-thought and think step by step.
    )
    }
    \end{AIbox}
    \caption{The template prompt for ranking summaries according to their performance on the errors described in \Cref{sec:errors}.}
    \label{fig:llm-ranking}
\end{figure*}

%% file: text/E_AdditionalContentRefinement.tex
% \begin{table}
%     \centering
%     \small
%     \begin{tabular}{llcccc}
%     \toprule
%     TP & FP      & BS & R-1 & R-2 & RLS \\
%     \midrule
%     dir & essential     & 16.20 & 33.73 & 07.46 & 20.53 \\
%     dir & CoT           & 16.16 & 33.89 & 07.57 & 20.41 \\
%     dir & Cor           & 16.19 & 33.89 & 07.52 & 20.39 \\
%     dir & CoT+Cor       & 16.35 & 33.90 & 07.56 & 20.58 \\
%     dir & Tra           & 15.28 & 33.89 & 07.82 & 20.99 \\
%     dir & Tra+CoT+Cor   & 15.12 & 33.78 & 07.94 & 21.31\\
%     \midrule
%     con & essential     & 14.27 & 29.79 & 05.58 & 18.26 \\
%     con & CoT           & 14.28 & 29.36 & 05.43 & 18.12 \\
%     con & Cor           & 15.11 & 29.64 & 05.55 & 18.37 \\
%     con & CoT+Cor       & 15.15 & 29.71 & 05.71 & 18.13 \\
%     con & Tra           & 14.96 & 29.90 & 05.55 & 18.55 \\
%     con & Tra+CoT+Cor   & 14.98 & 30.07 & 05.76 & 18.47 \\
%     \bottomrule
%     \end{tabular}
%     \caption{Score of the established evaluation metrics BERTScore (BS) and ROUGE (R-1 = ROUGE 1, R2 = ROUGE 2, RLS = ROUGE LSum).}
%     \label{tab:automatic_metric_scores_refinement}
% \end{table}

\Cref{tab:automatic_metric_scores_refinement} reports the BERTScore \cite{ZhangKWW20b} (re-weighted) and ROUGE \cite{Lin04b} scores for different combinations of FP and TP.
Note that the scores are very close to each other with slight variation, that does not allow for a thorough analysis.

%% file: text/F_Correction_CoT_contradiction.tex
\Cref{fig:confusion_CoT_Corr} demonstrates a case of contradicting information in CoT explanation and correction suggestion.

\label{sec:correction_cot_contradiction}
\begin{figure*}[t]
    \begin{AIbox}{Contradiction of Correction and CoT.}
    \parbox[t]{\textwidth}{
        \textbf{Chain-of-Thought:} \newline 
        1. The transcript discusses various aspects of the remote control design, including materials, shape, button placement, and technological features like speech recognition. \newline
        2. The predicted summary mentions a decision about the LCD display and back-lights, which is not found in the transcript. \newline
        3. The summary also talks about a scroll wheel on the side, which is mentioned in the transcript but not as a finalized decision. \newline
        4. The summary does not reflect the exploratory and undecided nature of the meeting's discussion on design elements. \newline
        5. The summary \hl{introduces specific design elements as decisions, which is not supported by the transcript}, indicating an extrinsic hallucination.
        \newline
        \newline
        \textbf{Correction:} \newline
        The summary should reflect the actual content of the meeting, focusing on the points that were discussed such as the importance of a fancy look-and-feel, the use of a scroll wheel for volume control, the consideration of technological innovations like speech recognition, and the overall design direction towards a modern, \hl{possibly iPod-like appearance}. Specific decisions about the LCD display and button placement should not be included unless they were explicitly mentioned in the transcript.
    }
    \end{AIbox}
    \caption{Confusion between CoT content and Correction suggestion.}
    \label{fig:confusion_CoT_Corr}
\end{figure*}